\pdfoutput=1

\documentclass[11pt]{article}

\usepackage{EMNLP2023}
\usepackage{booktabs}
\usepackage{amsmath}

\usepackage{graphicx}
\graphicspath{ {./img/} }

\usepackage{times}
\usepackage{latexsym}

\usepackage[T1]{fontenc}

\usepackage[utf8]{inputenc}

\usepackage{microtype}

\usepackage{inconsolata}

\title{Samsung R\&D Institute Philippines at WMT 2023}

\author{Jan Christian Blaise Cruz \\
Samsung R\&D Institute Philippines \\
\texttt{jcb.cruz@samsung.com}}

\begin{document}
\maketitle
\begin{abstract}
In this paper, we describe the constrained MT systems submitted by Samsung R\&D Institute Philippines to the WMT 2023 General Translation Task for two directions: en$\rightarrow$he and he$\rightarrow$en. Our systems comprise of Transformer-based sequence-to-sequence models that are trained with a mix of best practices: comprehensive data preprocessing pipelines, synthetic backtranslated data, and the use of noisy channel reranking during online decoding. Our models perform comparably to, and sometimes outperform, strong baseline unconstrained systems such as \textbf{\texttt{mBART50 M2M}} and \textbf{\texttt{NLLB 200 MoE}} despite having significantly fewer parameters on two public benchmarks: FLORES-200 and NTREX-128. 
\end{abstract}

\section{Introduction}
This paper describes Samsung R\&D Institute Philippines's submission to the WMT 2023 General Translation task. We participate in two translation directions: en$\rightarrow$he and he$\rightarrow$en, submitting two \textbf{constrained} single-direction models based on the Transformer \cite{vaswani2017attention} sequence-to-sequence architecture. We employ a number of best practices, using a comprehensive data preprocessing pipeline to ensure parallel data quality, create synthetic data through carefully-curated backtranslation, and use reranking methods to select the best candidate translations.

Our systems achieve strong performance on public benchmarks: 44.24 BLEU and 33.77 BLEU for FLORES-200 and NTREX-128 en$\rightarrow$he, respectively, and; 42.42 BLEU and 36.89 BLEU on FLORES-200 and NTREX-128 he$\rightarrow$en, respectively. Our systems outperform \textbf{\texttt{mBART50 M2M}} and slightly underperform against \textbf{\texttt{NLLB 200 MoE}} despite having significantly less parameters compared to these unconstrained baselines.

We detail our data preprocessing, model training, data augmentation, and translation methodology. Additionally, we illustrate hyperparameter sweeping setups and study the effects of hyperparameters during online decoding with reranking.

\section{Methodology}

\subsection{Data Preprocessing}

\begin{table*}[]
\centering
\begin{tabular}{l|rrr}
\hline
\multicolumn{1}{r|}{}    & \multicolumn{1}{c}{\textbf{Pairs}} & \multicolumn{1}{c}{\textbf{Words (en)}} & \multicolumn{1}{c}{\textbf{Words (he)}} \\ \hline
Original                 & 72,459,348                         & 701,991,594                             & 566,555,530                             \\
Original Filtered                 & 48,278,395                         & 385,975,984                             & 312,639,617                             \\
Synthetic en$\rightarrow$he          & 10,000,000                         & 165,595,289                             & 145,849,940                             \\
Synthetic en$\rightarrow$he Filtered & 7,143,725                          & 115,239,312                             & 95,954,020                              \\
Synthetic he$\rightarrow$en          & 73,278,018                         & 1,471,827,973                           & 1,056,677,671                           \\
Synthetic he$\rightarrow$en Filtered & 47,372,416                         & 659,409,236                             & 541,376,459                             \\ \hline
\end{tabular}

\caption{Corpus Statistics. ``Filtered'' refers to the number of pairs / words that remain after the filtering script is applied to the dataset. Note that ``Words'' is an approximation gathered by using the \texttt{wc -l *} command on the plaintext files.}
\label{tab:corpus-statistics}
\end{table*}

Given that a significant portion of the training dataset is synthetically-aligned, we need to use a comprehensive data preprocessing pipeline to ensure good translation quality. In particular, we use a combination of heuristic-based, ratio-based, and embedding-based methods to filter our data.

\paragraph{Heuristic-based} The following heuristic-based filters based on \citet{cruz2021improving} are used before applying the others:

\begin{itemize}
    \item \textbf{Language Filter} -- We use use \texttt{pycld3}\footnote{https://pypi.org/project/pycld2/} to filter out sentence pairs where one or both sentences have more than 30\% tokens that are neither English nor Hebrew.
    \item \textbf{Named Entity Filter} -- We use NER models \cite{bareket2021nemo,yang2018ncrf} to check if both sentences in a pair have matching entities (if any). Pairs that contain entities that do not match are removed. 
    \item \textbf{Numerical Filter} -- If one sentence in a pair has a number (ordinal, date, etc.), we also check the other sentence if a matching number is present. If a match is not detected, the pair is removed.
\end{itemize}

\paragraph{Ratio-based} We employ ratio-based filters on tokenized sentence pairs following \citet{cruz2022samsung} and \citet{sutawika2021data}. We first tokenize using SacreMoses\footnote{https://github.com/alvations/sacremoses} then apply  the following ratio-based filters:

\begin{itemize}
    \item \textbf{Length Filter} -- We remove pairs containing sentences with more than 140 characters.
    \item \textbf{Token Length Filter} -- We remove pairs that contain sentences with tokens that are more than 40 characters long.
    \item \textbf{Character to Token Ratio} -- We remove pairs where the ratio between character count and token count in at least one sentence is greater than 12.
    \item \textbf{Pair Token Ratio} -- We remove pairs where the ratio of tokens between the source and target sentences is greater than 4.
    \item \textbf{Pair Length Ratio} -- We remove pairs where the ratio between the string lengths of the source and target sentences is greater than 6.
\end{itemize}

\paragraph{Embedding-based} Finally, we experiment with the use of sentence embedding models to compute embedding-based similarity between a sentence pair. We use LaBSE \cite{feng2020language} models to embed both the source and target sentences then compute a cosine similarity score between the two. The pair must have a similarity score $0.7 \leq s \leq 0.96$ to be kept.

Statistics on the original and filtered corpus can be found on Table \ref{tab:corpus-statistics}.

After preprocessing the parallel data, we learn a shared BPE \cite{sennrich2015neural} vocabulary using SentencePiece\footnote{https://github.com/google/sentencepiece} \cite{kudo2018sentencepiece} with 32,000 units. All models in this paper use the same shared vocabulary.

\subsection{Model Architecture}
We experiment with two model sizes for each language pair: a \textbf{\texttt{Base}} model with 65M parameters and a \textbf{\texttt{Large}} model with 200M parameters. Both models use the standard Transformer \cite{vaswani2017attention} sequence-to-sequence architecture and are trained using Fairseq \cite{ott2019fairseq} with the hyperparameters listed in Table \ref{tab:training_hyperparams}. 

We parallelize with 8 NVIDIA Tesla P100 GPUs and initially train for a total of 100K steps for experimentation. For the submitted systems trained with backtranslated data, we train for a total of 1M steps.

\begin{table}[]
\centering
\begin{tabular}{lr}
\hline
\multicolumn{2}{c}{\textbf{Training Hyperparameters}}                \\ \hline
\multicolumn{1}{l|}{Parameters}   & \multicolumn{1}{r}{65M and 200M} \\
\multicolumn{1}{l|}{Vocab Size}   & 32,000                            \\
\multicolumn{1}{l|}{Tied Weights} & \multicolumn{1}{r}{Yes}          \\
\multicolumn{1}{l|}{Dropout}      & 0.3                             \\
\multicolumn{1}{l|}{Attention Dropout} & 0.1                             \\
\multicolumn{1}{l|}{Weight Decay} & 0.0                            \\
\multicolumn{1}{l|}{Label Smoothing} & 0.1                            \\
\multicolumn{1}{l|}{Optimizer}    & \multicolumn{1}{r}{Adam}         \\
\multicolumn{1}{l|}{Adam Betas}           & $\beta_1$=0.90, $\beta_2$=0.98       \\
\multicolumn{1}{l|}{Adam $\epsilon$}     & $\epsilon$=1e-6                             \\
\multicolumn{1}{l|}{Learning Rate}           & 7e-4                            \\
\multicolumn{1}{l|}{Warmup Steps} & 4,000                         \\
\multicolumn{1}{l|}{Total Steps}  & 1,000,000                     \\
\multicolumn{1}{l|}{Batch size}   & \multicolumn{1}{r}{64,000 tokens} \\ \hline
\end{tabular}
\caption{Hyperparameters used during training. When reporting model sizes, \textbf{\texttt{Base}} refers to 65M parameters, while \textbf{\texttt{Large}} refers to 200M.}
\label{tab:training_hyperparams}
\end{table}

\subsection{Backtranslation}
We use backtranslation \cite{sennrich2015improving} as a form of data augmentation to improve our initial models. We generate synthetic data via combined top-k and nucleus sampling:
\begin{equation}
    \begin{split}
         \sum_{i=0}^{\delta_{k}} P(\hat{y}^{(T)}_i | x; \hat{y}^{(T-1)}) * \delta_{temp} \leq \delta_{p}
    \end{split}
\end{equation}
where $\delta_{k}$ is the top values considered for top-k sampling, $\delta_{temp}$ is the temperature hyperparameter, and $\delta_{p}$ is the maximum total probability for nucleus sampling.

Backtranslation is only performed once using the provided monolingual data. We produce a total of 10,000,000 synthetic sentences for the en$\rightarrow$he direction and 73,278,018 synthetic sentences for the he$\rightarrow$en direction. The same data preprocessing used on the original parallel corpus is then applied to the synthetic corpus. We produce backtranslations using \textbf{\texttt{Large}} 100K models with the sampling hyperparameters  listed in Table \ref{tab:backtranslation_hyperparams}.

Statistics on generated synthetic data before and after filtering can be found on Table \ref{tab:corpus-statistics}.

\begin{table}[]
\centering
\begin{tabular}{lr}
\hline
\multicolumn{2}{c}{\textbf{Backtranslation Hyperparameters}}                \\ 
\hline
\multicolumn{1}{l|}{Top-k ($\delta_k$)}          & 50 \\
\multicolumn{1}{l|}{Top-p ($\delta_p$)}          & 0.93                           \\
\multicolumn{1}{l|}{Temperature ($\delta_t$)}    & 0.7          \\
\multicolumn{1}{l|}{Beam}           & 1.0                            \\
\multicolumn{1}{l|}{Length Penalty} & 1.0                             \\
\hline
\end{tabular}
\caption{Hyperparameters used during backtranslation.}
\label{tab:backtranslation_hyperparams}
\end{table}

\subsection{Noisy Channel Reranking}
We further improve translations by using Noisy Channel Reranking \cite{yee2019simple}, which reranks every candidate translation token $\hat{y}^{(T)}_i$ using Bayes' Rule, as follows:
\begin{equation}
    \begin{split}
        P(\hat{y}^{(T)}_i | x; \hat{y}^{(T-1)}) = \\
        \frac{P(x|\hat{y}^{(T-1)}) P(\hat{y}^{(T-1)})}{P(x)}
    \end{split}
\end{equation}
where $P(\hat{y}^{(T)}_i)$ refers to the probability of the $i$th candidate token at timestep $T$ given source sentence $x$ and current translated tokens $\hat{y}^{(T-1)}$. 

All probabilities are parameterized as standard encode-decoder Transformer neural networks: the \textbf{Direct Model} $f_{\phi_{D}}(x,\hat{y}^{(T-1)})$ models $P(\hat{y}^{(T)}_i | x; \hat{y}^{(T-1)})$ or translation between source to target language; the \textbf{Channel Model} $f_{\phi_{C}}(x|\hat{y}^{(T-1)})$ models $P(x|\hat{y}^{(T-1)})$, or the probability of the target translating back into the predicted translation, and; the \textbf{Language Model} $f_{\phi_{L}}(\hat{y}^{(T-1)})$ models $P(\hat{y}^{(T-1)})$ or the probability of the translated sentence to exist. $P(x)$ is generally not modeled since it is constant for all $y$. This allows us to leverage a strong language model to guide the outputs of the direct model, while using a channel model to constrain the preferred outputs of the language model (which may be unrelated to the source sentence).

During beam search decoding, we rescore the top candidates using the following linear combination of all three models:
\begin{equation}
    \begin{split}
        P(\hat{y}^{(T)}_i | x; \hat{y}^{(T-1)})^{'} = \frac{1}{t} log(P(x|\hat{y}^{(T-1)}) \\
        + \frac{1}{s} [ \delta_{ch} log(P(x|\hat{y}^{(T-1)}) \\
        + \delta_{lm} log(P(\hat{y}^{(T-1)})) ]
    \end{split}
\end{equation}
where $s$ and $t$ are source / target debiasing terms, $\delta_{ch}$ refers to the weight of the channel model, and $\delta_{lm}$ refers to the weight of the language model.

For Noisy Channel Reranking, our direct and channel models use the same size and setup at all times (i.e. if the direct model is a \textbf{\texttt{Large}} model trained for 100K steps, then the channel model is also a \textbf{\texttt{Large}} model trained for 100K steps in the opposite translation direction). 

For the language model, we train one \textbf{\texttt{Base}}-sized decoder-only Transformer language model for English and one for Hebrew. We concatenate the cleaned data from the parallel corpus with the provided monolingual data for each language to train the LM. We use the same training setup as with translation models, except we use a weight decay of 0.01 and a learning rate of 5e-4.

Hyperparameters used for decoding with Noisy Channel Reranking can be found in Table \ref{tab:decoding_hyperparams}. 

\begin{table}[]
\centering
\begin{tabular}{lr}
\hline
\multicolumn{2}{c}{\textbf{Decoding Hyperparameters}}                \\ 
\hline
\multicolumn{1}{l|}{Beam}                           & 5 \\
\multicolumn{1}{l|}{Length Penalty}                 & 1.0                           \\
\multicolumn{1}{l|}{k2}                             & 5          \\
\multicolumn{1}{l|}{CM Top-k}                       & 500                            \\
\multicolumn{1}{l|}{$\delta_{ch}$ en$\rightarrow$he}  & 0.2297                             \\
\multicolumn{1}{l|}{$\delta_{lm}$ en$\rightarrow$he}  & 0.2056          \\
\multicolumn{1}{l|}{$\delta_{ch}$ he$\rightarrow$en}  & 0.2998                            \\
\multicolumn{1}{l|}{$\delta_{lm}$ he$\rightarrow$en}  & 0.2594                             \\
\hline
\end{tabular}
\caption{Hyperparameters used for the final submission models. The values listed for $\delta_{ch}$ and $\delta_{lm}$ are the ones used for the final submission models. For testing with \textbf{\texttt{Large} 100K} models, we set both $\delta_{ch}$ and $\delta_{lm}$ to 0.3. ``k2'' refers to the number of candidates sampled per beam while ``CM Top-k'' refers to the number of most frequent tokens in the channel model's vocabulary that is used as its output vocabulary during decoding to save space.}
\label{tab:decoding_hyperparams}
\end{table}

\subsection{Evaluation}
We evaluate our models using two metrics: BLEU \cite{papineni2002bleu} and ChrF++ \cite{popovic2015chrf}, both scored via SacreBLEU\footnote{SacreBLEU outputs the following signature for evaluation: nrefs:1|case:mixed|eff:no|tok:spm-flores|smooth:exp|version:2.2.1} \cite{post-2018-call}. We develop our models using both the FLORES 200 \cite{costa2022no} and NTREX 128 \cite{federmann-etal-2022-ntrex} datasets, using the validation sets during training and reporting scores on the test sets.

To benchmark our models' performance, we mainly compare BLEU and ChrF++ against two (unconstrained) models: \textbf{\texttt{mBART 50 M2M}} \cite{tang2020multilingual}, a ~610M-parameter finetuned version of \textbf{\texttt{mBART}} for many-to-many translation, and \textbf{\texttt{NLLB 200 MoE}} \cite{costa2022no}, the full 54.5B-parameter mixture-of-experts version of \textbf{\texttt{NLLB 200}} for many-to-many translation.

\subsection{Hyperparameter Search}
To find the best values for $\delta_{ch}$ and $\delta_{lm}$, as well as to understand how these parameters affect performance, we use Bayesian Hyperparameter Search. We use the \textbf{\texttt{Large}} 1M + BT models and run 1000 iterations of search, keeping the length penalty static at 1.0, and sampling both $\delta_{ch}$ and $\delta_{lm}$ from a gaussian with minimum of 0.01 and maximum of 0.99.

We perform this for both en$\rightarrow$he and he$\rightarrow$en translation directions and use the results for the final submission model.

\section{Results}

A summary of our results on benchmarks can be found on Table \ref{tab:results}.

\begin{table*}[]
\centering
\resizebox{\textwidth}{!} {
\begin{tabular}{l|rrrr|rrrr|}
\cline{2-9}
                                                  & \multicolumn{4}{c|}{\textbf{FLORES-200}}                                                                                                         & \multicolumn{4}{c|}{\textbf{NTREX-128}}                                                                                                          \\ \cline{2-9} 
                                                  & \multicolumn{2}{c|}{\textbf{EN $\rightarrow$ HE}}                        & \multicolumn{2}{c|}{\textbf{HE $\rightarrow$ EN}}                       & \multicolumn{2}{c|}{\textbf{EN $\rightarrow$ HE}}                        & \multicolumn{2}{c|}{\textbf{HE $\rightarrow$ EN}}                       \\ \hline
\multicolumn{1}{|c|}{\textbf{Model}}              & \multicolumn{1}{c}{\textbf{BLEU}} & \multicolumn{1}{c|}{\textbf{ChrF++}}  & \multicolumn{1}{c}{\textbf{BLEU}} & \multicolumn{1}{c|}{\textbf{ChrF++}} & \multicolumn{1}{c}{\textbf{BLEU}} & \multicolumn{1}{c|}{\textbf{ChrF++}}  & \multicolumn{1}{c}{\textbf{BLEU}} & \multicolumn{1}{c|}{\textbf{ChrF++}} \\ \hline
\multicolumn{1}{|l|}{\textbf{\texttt{Base}} 100K}                   & 39.88                             & \multicolumn{1}{r|}{56.34}          & 12.06                             & 29.46                              & 31.47                             & \multicolumn{1}{r|}{48.32}          & 29.85                             & 52.53                              \\
\multicolumn{1}{|l|}{\textbf{\texttt{Base}} 100K + NC}              & 40.22                             & \multicolumn{1}{r|}{56.55}          & 38.75                             & 60.52                              & 32.10                             & \multicolumn{1}{r|}{48.93}          & 31.86                             & 54.57                              \\
\multicolumn{1}{|l|}{\textbf{\texttt{Base}} 100K + BT}              & 41.50                             & \multicolumn{1}{r|}{57.46}          & 38.73                             & 60.80                              & 31.27                             & \multicolumn{1}{r|}{47.90}          & 34.09                             & 56.10                              \\
\multicolumn{1}{|l|}{\textbf{\texttt{Base}} 100K + BT + NC}         & 41.66                             & \multicolumn{1}{r|}{57.59}          & 40.43                             & 62.17                              & 32.05                             & \multicolumn{1}{r|}{48.62}          & 35.76                             & 57.65                              \\
\multicolumn{1}{|l|}{\textbf{\texttt{Large}} 100K}                  & 41.26                             & \multicolumn{1}{r|}{57.46}          & 39.07                             & 60.06                              & 32.49                             & \multicolumn{1}{r|}{48.95}          & 31.08                             & 53.19                              \\
\multicolumn{1}{|l|}{\textbf{\texttt{Large}} 100K + NC}             & 41.46                             & \multicolumn{1}{r|}{57.64}          & 40.53                             & 61.49                              & 32.80                             & \multicolumn{1}{r|}{49.34}          & 33.12                             & 55.16                              \\
\multicolumn{1}{|l|}{\textbf{\texttt{Large}} 100K + BT}             & 43.32                             & \multicolumn{1}{r|}{58.62}          & 40.91                             & 61.58                              & 32.90                             & \multicolumn{1}{r|}{49.11}          & 35.48                             & 56.04                              \\
\multicolumn{1}{|l|}{\textbf{\texttt{Large}} 100K + BT + NC}        & 43.26                             & \multicolumn{1}{r|}{58.72}          & 41.92                             & 62.64                              & 33.18                             & \multicolumn{1}{r|}{49.42}          & 36.79                             & 57.37                              \\
\multicolumn{1}{|l|}{\textbf{\texttt{Large}} 1M + BT}               & 43.76                             & \multicolumn{1}{r|}{58.29}          & 41.00                             & 61.16                              & 33.35                             & \multicolumn{1}{r|}{49.22}          & 35.83                             & 56.02                              \\
\multicolumn{1}{|l|}{\textbf{\textbf{\texttt{Large}} 1M + BT + NC}} & \textbf{44.24}                    & \multicolumn{1}{r|}{\textbf{59.36}} & \textbf{42.42}                    & \textbf{62.21}                     & \textbf{33.77}                    & \multicolumn{1}{r|}{\textbf{49.69}} & \textbf{36.89}                    & \textbf{56.92}                     \\ \hline
\multicolumn{1}{|l|}{\textbf{\texttt{mBART50 M2M}} (610M)}               & 19.49                             & \multicolumn{1}{r|}{46.7}          & 30.50                             & 55.00                              & 14.80                             & \multicolumn{1}{r|}{42.30}          & 27.02                             & 51.21                              \\
\multicolumn{1}{|l|}{\textbf{\texttt{NLLB 200 MoE}} (54.5B)}               & 46.80                             & \multicolumn{1}{r|}{59.80}          & 49.00                             & 67.40                              & -                             & \multicolumn{1}{r|}{-}          & -                             & -                              \\
\hline
\end{tabular} }
\caption{Compiled results for all experiments. ``BT'' refers to the model being trained with backtranslated data in addition to original filtered data. ``NC'' refers to the use of Noisy Channel Reranking. Evaluation scores for \textbf{\texttt{NLLB 200 MoE}} are taken from its official published scores for FLORES-200. We fail to report independent NTREX-128 scores for \textbf{\texttt{NLLB 200 MoE}} due to a lack of computational resources.}
\label{tab:results}
\end{table*}

\begin{figure*}
    \centering
    \includegraphics[width=\textwidth]{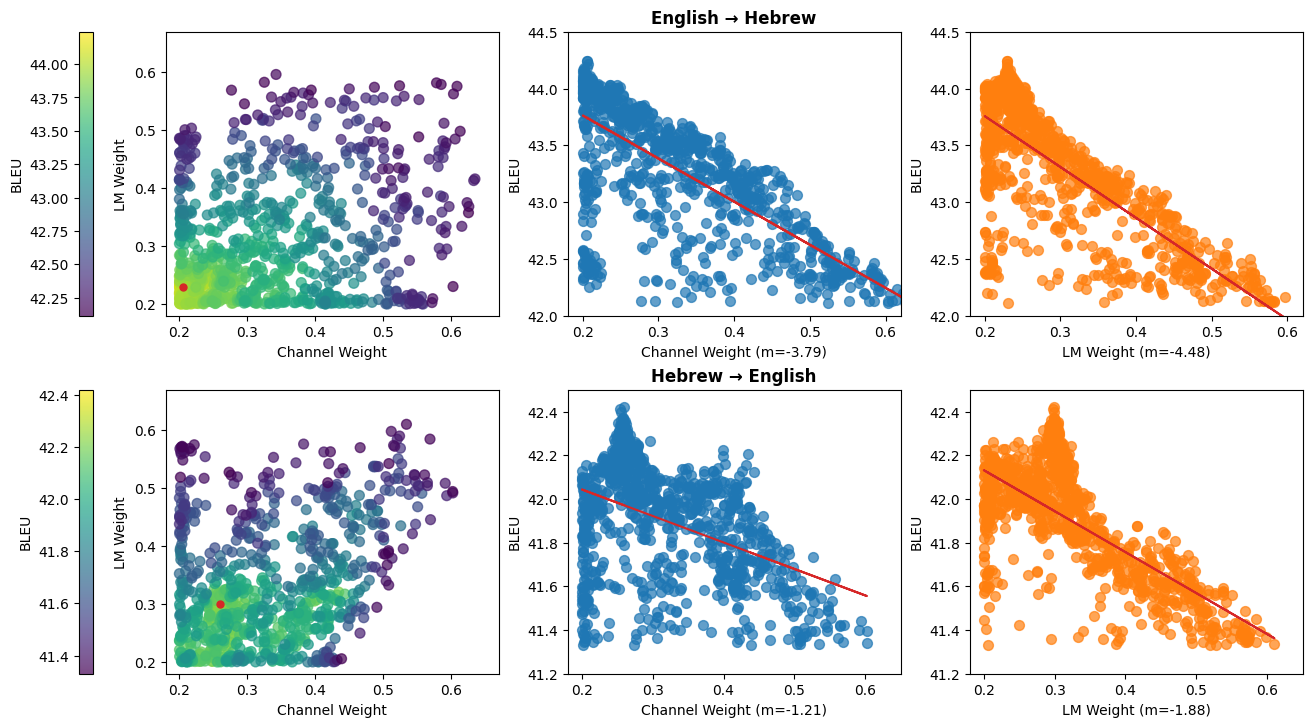}
    \caption{Bayesian hyperparameter search results for $\delta_{ch}$ and $\delta_{lm}$ while keeping constant length penalty. The leftmost column shows BLEU score against both $\delta_{ch}$ and $\delta_{lm}$ with the best performing model (\textbf{\texttt{Large}} 1M + BT + NC) plotted in red. The middle and rightmost columns show $\delta_{ch}$ and $\delta_{lm}$ against BLEU, respectively, with their respective regression lines (in red) and regression coefficient (m) in the caption.}
    \label{fig:noisy_channel_param}
\end{figure*}

\subsection{Benchmarking Results}

Our submission systems (\textbf{\texttt{Large} 1M + BT + NC}) exhibit strong performance on both translation directions. On FLORES-200, we achieve 44.24 BLEU for en$\rightarrow$he and 42.42 BLEU for he$\rightarrow$en. The same systems score 33.77 BLEU for en$\rightarrow$he and 36.89 BLEU for he$\rightarrow$en on NTREX-128. 

We note that these systems perform strongly when compared against much larger, unconstrained baseline models. On FLORES-200, we significantly outperform \textbf{\texttt{mBART 50 M2M}} on en$\rightarrow$he by +24.75 BLEU and on he$\rightarrow$en by +11.92 BLEU despite having 67\% less parameters (200M vs 610M). Notably, our system performs only slightly worse compared to \textbf{\texttt{NLLB 200 MoE}} despite having \textbf{96\% less parameters} compared to the mixture-of-experts model. On FLORES-200, we perform -2.56 BLEU worse on en$\rightarrow$he and -6.58 BLEU worse on he$\rightarrow$en compared to \textbf{\texttt{NLLB 200 MoE}}.


\subsection{Hyperparameter Search Results}

In order to find optimal hyperparameters for both $\delta_{ch}$ and $\delta_{lm}$, we ran bayesian hyperparameter search for both at the same time while keeping length penalty static. We plot the results of the hyperparameter search over 1000 iterations in Figure \ref{fig:noisy_channel_param}.

We observe that performance is optimal when both hyperparameters are set to 0.2$\sim$0.3, making performance increasingly worse as both hyperparameters approach closer to 1. We hypothesize that this signifies the model capturing the original distribution close enough that it does not need much correction or aid from the accompanying language model. Noisy channel reranking, however, is still empirically shown to be useful in this case as guidance from the language model produces better candidates in cases where the direct model may be searching a too-constrained space.

\subsection{Ablations}
We explored multiple configurations of our submission systems in terms of model size, presence of synthetic data during training, and the use of reranking methods during online decoding. Our results show that each step improves performance directly:
\begin{itemize}
    \item The initial \textbf{\texttt{Base} 100K} performs at 39.88 BLEU for en$\rightarrow$he on FLORES-200.
    \item Increasing the size to 200M parameters (\textbf{\texttt{Large} 100K}) improves performance by +1.38 BLEU.
    \item Adding backtranslated data (\textbf{\texttt{Large} 100K + BT}) is by far the most beneficial, improving performance by +2.06 BLEU.
    \item We then experiment with longer training times (1M iterations for \textbf{\texttt{Large} 1M + BT}) to adapt to the new dataset size, increasing the score by +0.44 BLEU.
    \item Finally, using noisy channel reranking (\textbf{\texttt{Large} 1M + BT + NC}) improves the score by +0.48 BLEU.
\end{itemize}

Overall, all of our methods improve performance by a total of 4.36 BLEU for the en$\rightarrow$he direction on FLORES-200. 

We note an interesting jump in performance from \textbf{\texttt{Base} 100K} to \textbf{\texttt{Large} 1M + BT + NC} on the FLORES-200 he$\rightarrow$en direction at +30.36 BLEU. \textbf{\texttt{Base} 100K} underperforms at 12.06 BLEU, and we hypothesize that this is due to the model not having enough capacity to embed information from Hebrew, which causes it to greatly benefit from the guidance of a language model during noisy channel reranking.

\section{Conclusion}
In this paper, we describe our submissions to the WMT 2023 General Translation Task. We participate in two constrained tracks: en$\rightarrow$he and he$\rightarrow$en.

We submit two monodirectional models based on the Transformer architecture. Both models are trained using a mix of original and synthetic backtranslated data, filtered and curated using a comprehensive data processing pipeline that combines embedding-based, heuristic-based, and ratio-based filters. Additionally, we employ noisy channel reranking to improve translation candidates using a language model and a channel model trained in the opposite direction.

On two benchmark datasets, our systems outperform \textbf{\texttt{mBART50 M2M}} and perform slightly worse than \textbf{\texttt{NLLB 200 MoE}}, both unconstrained systems with significantly more parameters.

Our results show that established best practices still perform strongly on constrained systems without the need for extraneous data sources as is with unconstrained systems for the same translation directions.

\section*{Limitations}
We benchmark on datasets that are publicly available with permissive licenses for research. 

We note that we are unable to study scale properly for translation models due to a lack of stronger compute resources. The same constraint also prevents us from training multiple iterations of the same model with differing random seeds. Our systems' true performance may thus be higher or lower depending on the machine random state at the start of training time. 

Lastly, our models are trained on Hebrew, which is a language that we do not speak. We are therefore unable to manually evaluate if the output translations are correct, natural, or semantically sound.

\section*{Ethical Considerations}
Our paper replicates best practices in data preprocessing, model training, and online decoding for translation models. Within our study, we aim to create experiments that replicate prior work under comparable experimental conditions to ensure fairness in benchmarking.

Given that we do not speak the target language in the paper, we report performance in comparison to other existing models. We do not claim that ``strong'' performance in a computational setting correlates with good translations from a human perspective.

Lastly, while we do not use human annotators for this paper, the conference (WMT) itself does for human evaluations on the General Translation Task. We disclose this fact and note that annotations (and therefore scores) may be different across many speakers of Hebrew.

\bibliography{emnlp2023}
\bibliographystyle{acl_natbib}




\end{document}